\documentclass[nohyperref]{article}

\usepackage{microtype}
\usepackage{graphicx}
\usepackage{subfigure}
\usepackage{booktabs} % for professional tables
\usepackage{amsmath}

\usepackage{hyperref}

\DeclareMathOperator*{\argmax}{arg\,max}
\DeclareMathOperator*{\argmin}{arg\,min}

\usepackage[accepted]{icml2022}

\usepackage{amsmath}
\usepackage{bm}
\usepackage{amssymb}
\usepackage{mathtools}
\usepackage{amsthm}
\usepackage{float}

\usepackage[capitalize,noabbrev]{cleveref}

\theoremstyle{plain}

\theoremstyle{definition}

\theoremstyle{remark}

\usepackage[textsize=tiny]{todonotes}

\usepackage{enumitem}

\usepackage{xcolor}

\icmltitlerunning{Making Corgis Important for Honeycomb Classification: Adversarial Attacks on Concept-based Explainability Tools}

\begin{document}

\twocolumn[
\icmltitle{Making Corgis Important for Honeycomb Classification: Adversarial Attacks on Concept-based Explainability Tools}




\begin{icmlauthorlist}
\icmlauthor{Davis Brown}{yyy}
\icmlauthor{Henry Kvinge}{yyy,comp}
% \icmlauthor{Firstname3 Lastname3}{comp}
\end{icmlauthorlist}

\icmlaffiliation{yyy}{Pacific Northwest National Laboratory}
\icmlaffiliation{comp}{Department of Mathematics, University of Washington}
% \icmlaffiliation{sch}{Department of Mathematics, Colorado State University}

\icmlcorrespondingauthor{Davis Brown}{davis.brown@pnnl.gov}

\icmlkeywords{Explainability, adversarial attacks}

\vskip 0.3in
]



\printAffiliationsAndNotice{}  % leave blank if no need to mention equal contribution

\begin{abstract}
Methods for model explainability have become increasingly critical for testing the fairness and soundness of deep learning. Concept-based interpretability techniques, which use a small set of human-interpretable concept exemplars in order to measure the influence of a concept on a model's internal representation of input, are an important thread in this line of research. In this work we show that these explainability methods can suffer the same vulnerability to adversarial attacks as the models they are meant to analyze. We demonstrate this phenomenon on two well-known concept-based interpretability methods: TCAV and faceted feature visualization. We show that by carefully perturbing the examples of the concept that is being investigated, we can radically change the output of the interpretability method. The attacks that we propose can either induce positive interpretations (polka dots are an important concept for a model when classifying zebras) or negative interpretations (stripes are not an important factor in identifying images of a zebra). Our work highlights the fact that in safety-critical applications, there is need for security around not only the machine learning pipeline but also the model interpretation process. 
\end{abstract}

\section{Introduction}

Deep learning models have achieved superhuman performance in a range of activities from image recognition to complex games \citep{lecun2015deep,silver2017mastering}. Unfortunately, these gains have come at the expense of model interpretability, with massive, overparametrized models being used to achieve state-of-the-art results. This is a major limitation when deep learning is applied to domains such as healthcare \citep{miotto2018deep}, criminal justice \citep{li2018deep}, and finance \citep{huang2020deep}, where a prediction needs to be explainable to the user in order to be trusted. This has led to a surge of interest in tools that can illuminate the underlying decision making process of deep learning models.

Concept-based interpretability methods (CBIMs) are a family of explainability techniques that are increasingly popular. The critical observation underlying these methods is that in many scenarios, low-level statistics such as the importance of individual pixels in an input image (as provided by saliency methods for example), cannot deliver the depth of insight that a user needs in complex, real-world situations. CBIMs instead rely on a user provided collection of positive examples (tokens) of a human-interpretable concept which are then used to probe a model. CBIMs have now been successfully applied to a range of applications, from healthcare tasks \cite{graziani_regression_2018, 10.1145/3450439.3451858} to understanding the strategies of a deep learning-based chess engine \cite{mcgrath2021acquisition}. In this paper we focus on two examples of CBIMs that capture both the diversity and power of these methods: Testing with Concept Activation Vectors (TCAV) \citep{kim2018interpretability} and Faceted Feature Visualization (FFV) \citep{goh2021multimodal}.

Besides being inherently black-box in nature, deep learning models have also been shown to be vulnerable to adversarial attacks where small perturbations to model input result in dramatic changes to model output \citep{szegedy2013intriguing}. This phenomenon is concerning when deep learning tools are deployed in safety-critical environments. But if explainability methods are an important component in a machine learning system, then the robustness of these methods themselves is nearly as important as the robustness of the model. In this paper we explore the vulnerability of CBIMs to adversarial attacks. 

Our analysis identifies the small number of concept tokens used in CBIM methods as a single point of failure in the entire interpretability pipeline. Indeed, subtle changes to a few centralized tokens representing a concept could result in dramatic misinterpretation of many subsequent input. In the case where the reasoning behind a model's predictions is almost as important as the model's predictions themselves, this could result in a model being taken out of deployment. Despite the fact that CBIM methods can take a variety of forms, our proposed attack which we call a {\emph{token pushing (TP) attack}} is applicable to many of them since it targets the linear probe mechanism that is common to nearly all.

%We test TP attacks against two popular CBIMs: Testing with Concept Activation Vectors (TCAV) \citep{kim2018interpretability} and Faceted Feature Visualization (FFV) \citep{goh2021multimodal}. While TCAV and FFV are similar in that they are both concept-based, their output is quite different. TCAV quantifies the extent to which a concept is important to a model's prediction for a specific input dataset. A variant of this method was recently a central component of \citep{mcgrath2021acquisition}, which used it to provide evidence that models such as AlphaZero learn human chess concepts. FFV on the other hand, can be used to produce visualizations that represent how individual neurons capture a specific concept. We show that TP attacks are effective for both TCAV and FFV. For example, a TP attack causes TCAV to give output indicating that stripes are not an important feature to the class `zebra.' On the other hand, a TP attack can radically change the feature visualizations generated by FFV (Figure \ref{fig-ffv-visualizations}). 

We evaluate TP attacks against both TCAV and FFV on pretrained ImageNet models \citep{deng2009imagenet, marcel2010torchvision} using the Describable Textures Dataset \citep{cimpoi14describing} as a source of concept tokens and on models trained on Caltech-UCSD Birds 200 \cite{welinder2010caltech} using images with specific attributes as concept tokens. Through our experiments we show that, provided that it uses a linear probe, the TP attack does not even require the adversary to know what interpretability method is being used. The same perturbations that cause TCAV to fail also cause FFV to fail. Finally, our TP attack possesses moderate transferability between different model architectures, meaning that a TP attack can be developed via a surrogate model even when the defender model architecture is unknown.  

Our contributions in this paper include the following.
\begin{itemize}[noitemsep]
    \item Formalization of an adversarial threat model for post-hoc concept-based interpretability methods that identifies concept tokens as a single point of failure.
    \item Introduction of TP attacks which cause deliberate misinterpretation by disrupting the linear probe mechanism underlying many concept-based interpretability methods.
    \item Demonstration of the effectiveness of TP attacks on TCAV and FFV.
    \item Introduction of the first (to our knowledge) adversarial attack on feature visualization.
\end{itemize}

\section{TCAV and linear interpretability}
\label{sect-TCAV}

In this section we describe the method of testing with concept activation vectors (TCAV) \citep{kim2018interpretability}. TCAV has become a popular interpretability technique that has been used in a range of applications \cite{lucieri2020interpretability,janik2021interpretability,thakoor2020robust}. Let $f: X \rightarrow \mathbb{R}^d$ be a neural network which is composed of $n$ layers and designed for the task of classifying whether a given input $x \in X$ belongs to one of $d$ different classes. Write $f_\ell: X \rightarrow \mathbb{R}^{d_\ell}$ for the composition of the first $\ell$ layers so that $f_n = f$ and $d_n = d$ and let $h_\ell: \mathbb{R}^{d_\ell} \rightarrow \mathbb{R}^d$ be the composition of the last $n-\ell$ layers of the network so that $f = h_\ell \circ f_\ell$ for any $1 \leq \ell \leq n-1$. Let $C$ be a concept for which we have a set of positive examples (tokens) $P_C = \{x^C_i\}_i$ and negative examples $N_C = \{x^N_i\}_i$, both belonging to $X$. These are represented in the $\ell$th layer of $f$ as the points $f_\ell(P_C)$ and $f_\ell(N_C)$ respectively. One can apply a binary linear classifier to separate these two sets of points, resulting in a hyperplane in $\mathbb{R}^{d_\ell}$. We represent this hyperplane by the normal vector $v_C^\ell \in \mathbb{R}^{d_\ell}$ that points into the region corresponding to the points $f_\ell(P_C)$. $v_C^\ell$ is called the {\emph{concept activation vector}} in layer $\ell$ associated with concept $C$. One can think of $v_C^\ell$ as the vector that points toward $C$-ness in the $\ell$th layer of the network. 

Let $h_{\ell,k}$ denote the $k$th output coordinate of $h_\ell$ corresponding to class $k$. In the classification setting, $h_{\ell,k}$ then represents the model's confidence that input belongs to class $k$. To better understand the extent to which concept $C$ influences the model's confidence of $x \in X$ belonging to class $k$ we compute:
\begin{equation} \label{eq:tcav}
    S_{C, k, l} = \nabla h_{\ell, k}\left(f_{\ell}(x)\right) \cdot v_{C}^{l}.
\end{equation}
A positive value of $S_{C, k, l}$ indicates that increasing $C$-ness of $x$ makes the model more confident that $x$ belongs to class $k$. The {\emph{magnitude TCAV score}} for a dataset $D$ is defined as the sum of $S_{C, k, l}(x)$ over all $x \in D_k$, where $D_k$ is the subset of $D$ consisting of all instances predicted as belonging to class $k$, divided by $|D_k|$. We compare the TCAV magnitude of the positive concept images with the TCAV magnitude for random images in the layer, and use a standard two-sided $t$-test to test for significance. We can also compute the {\emph{relative TCAV score}}, which replaces the set of negative natural images in $N_C$ with images representing a specific concept. % \hjk{HK: Removing mention of experiments here.}

\subsection{Faceted Feature Visualization}

\cite{goh2021multimodal} introduced a new concept-based feature visualization objective for neuron-level interpretability, \emph{Faceted Feature Visualization (FFV)}. The objective disambiguates polysemantic neurons by imposing a prior towards a linear concept $C$ in the optimization objective. \citet{goh2021multimodal} also utilizes the linear probe framework with sets of positive and negative examples of a concept $C$ ($P_C$ and $N_C$ respectively). Similar to the TCAV method, one trains a binary linear classifier on $f_\ell(P_C)$ and $f_\ell(N_C)$ to obtain CAV $v_{C}^{l}$. To visualize output that tends to activate a neuron at layer $\ell$, position $i$, while at the same time steering the visualization toward a specific concept, the authors optimize for the objective function:
\begin{equation}\label{eq:ffv}
\argmax_{x \in X} f_{\ell,i}(x)+v_{C}^{l} \cdot (f_{\ell}(x) \odot \nabla f_{\ell,i}(x)),
\end{equation}
where $\odot$ is the Hadamard product. %Note that the first term helps find $x$ which result in a strong activation of $f_{\ell,i}$, while the second term finds $x$ such that $f_{\ell}(x)$ tends to point in the direction of $v_C^\ell$. 

\begin{figure*}[ht]
\centering
\includegraphics[width=4.7 in]{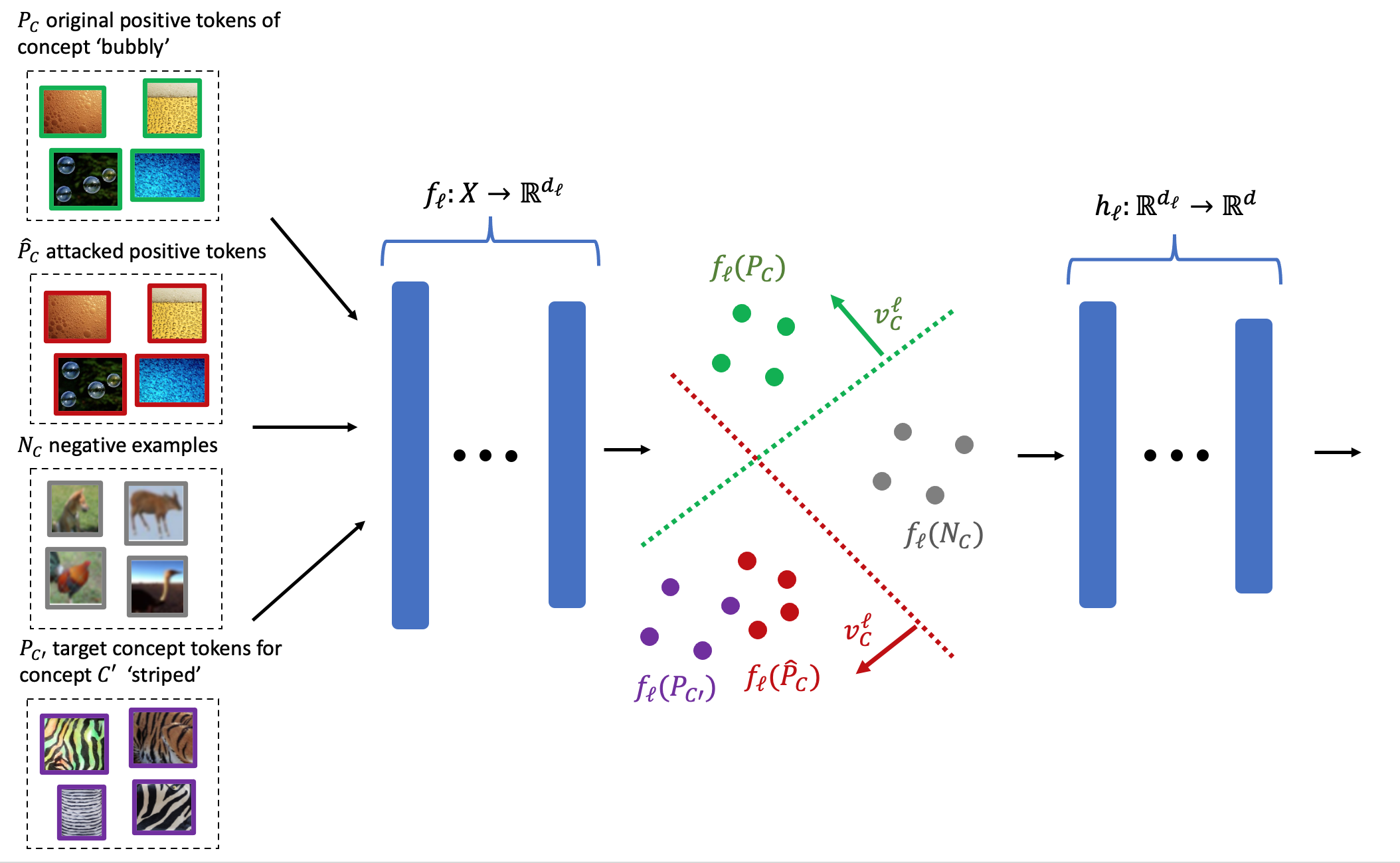}
\caption{A schematic of the targeted TP attack. $P_C$ is the original set of positive examples of concept `bubbly' $C$ (green), $N_C$ is the set of negative examples of concept $C$ (grey), $P_{C'}$ is the set of positive examples for target concept `striped' $C'$ (purple), and $\hat{P}_C$ is $P_C$ after being perturbed by the TP attack. When $P_C$ is perturbed to $\hat{P}_C$, it shifts CAV $v^\ell_C$ so that it is more closely aligned to the CAV for `striped'. The result is that input that is intended to be interpreted in terms of concept `bubbly' is actually interpreted with respect to the concept `striped'.}
\centering
\label{fig-schematic-of-attack} 
\end{figure*}

\begin{figure*}[ht]%
\centering
\label{fig:CUB}%
\includegraphics[height=1.5in]{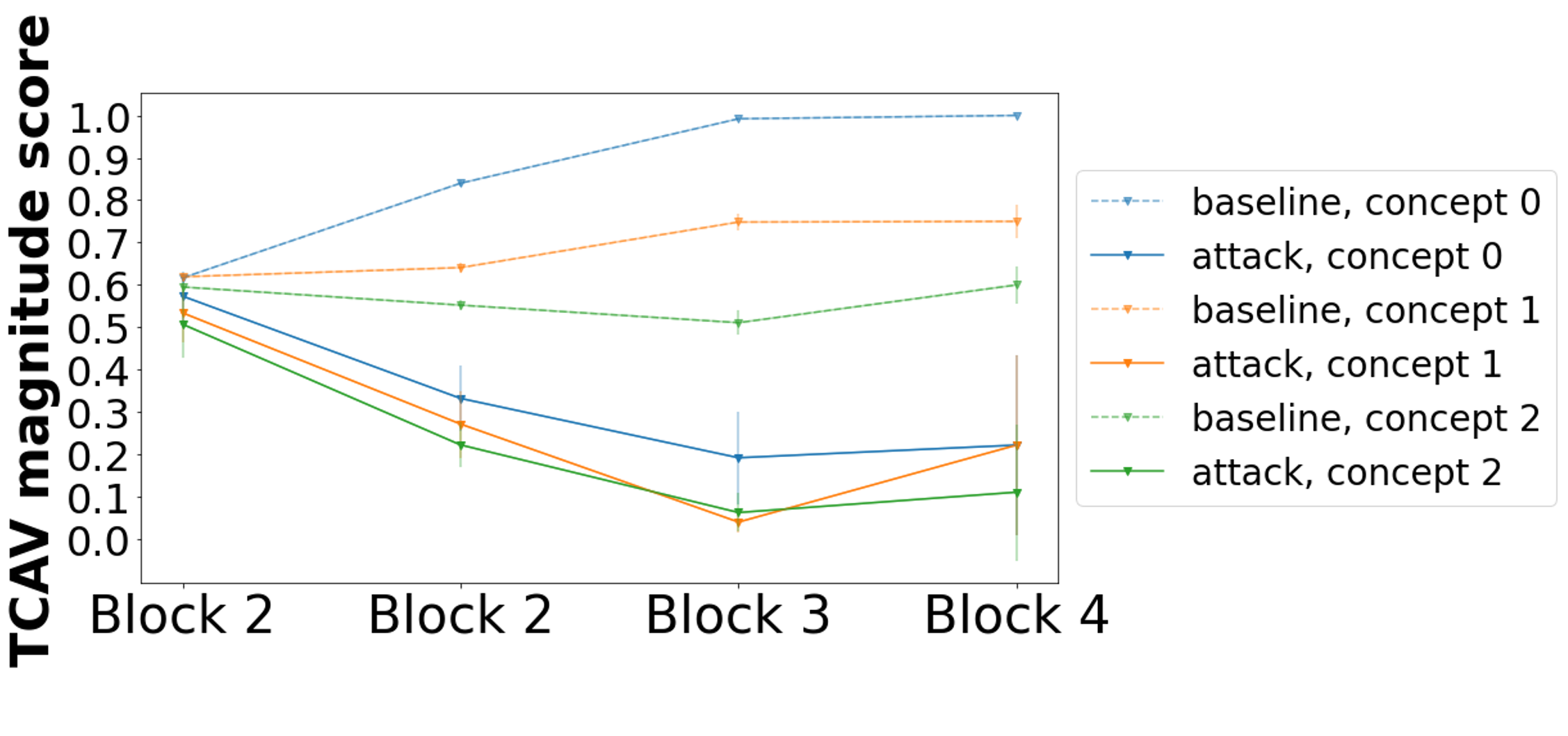}%
\includegraphics[height=1.5in]{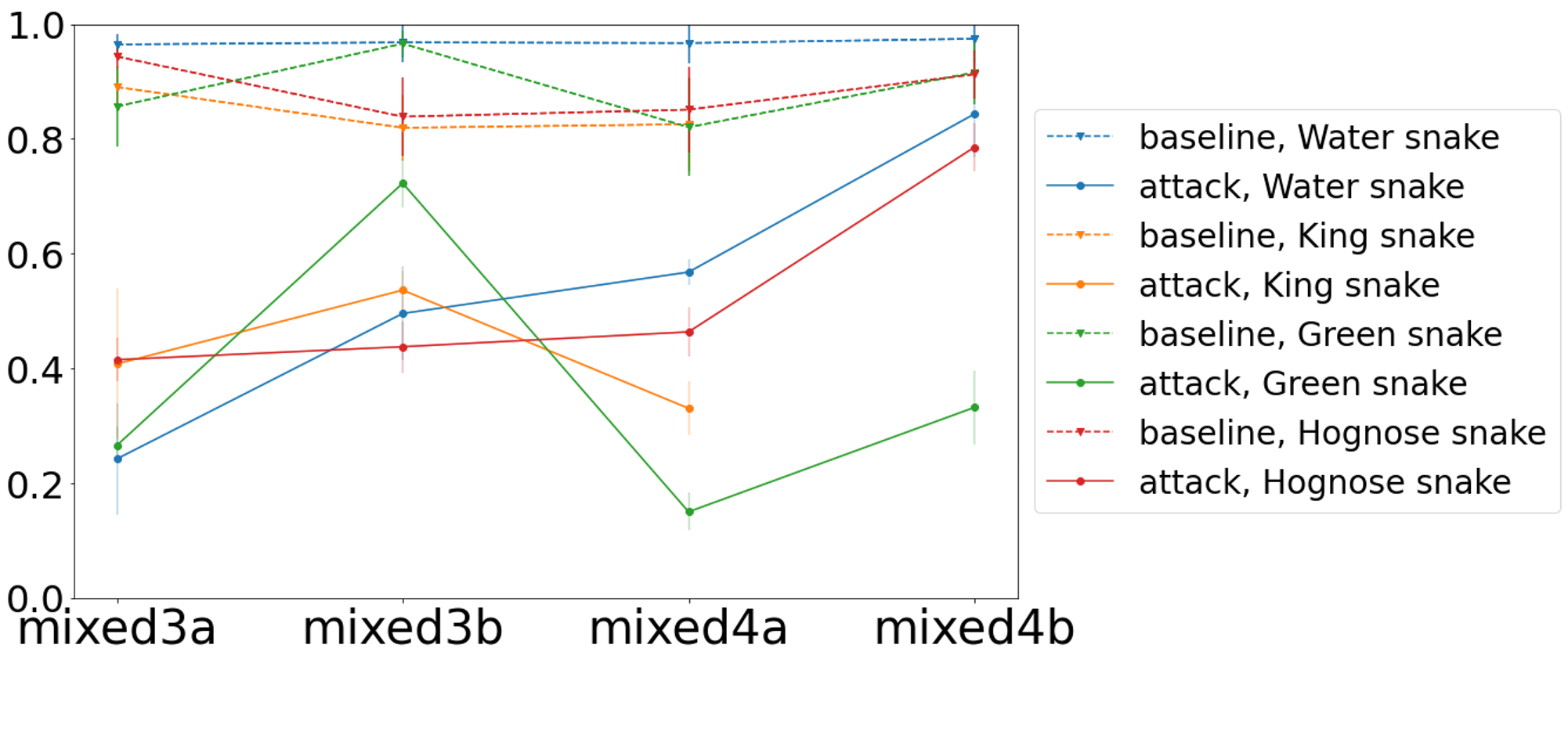}
\caption{The untargeted TP attack on three different concepts for a ResNet-18 trained on Caltech-UCSD Birds 200 with TCAV magnitude scores with respect to the class `brewer blackbird' (left) and the scaly DTD concept for an InceptionV1 trained on ImageNet with TCAV magnitude scores with respect to snake classes in ImageNet (righ). Note that the plot on the left varies the concepts but keeps the class, `brewer blackbird', fixed while and plot on the right varies the class while keeping the concept, `scaly', fixed.}
\end{figure*}

\section{An Attack on the Tokens of Concept}
\label{sect-tokens}

Traditionally, an adversarial attack \citep{szegedy2013intriguing} on a model $f$ is a small perturbation $\delta$ that, when applied to a specific input $x$, results in large changes to model prediction $f(x)$. The meaning of `small' is usually specified by a metric such as an $\ell_p$-norm and can either be a hard or soft constraint. In this work we use projected gradient descent (PGD) \citep{madry2018towards} to construct our attacks, since it is widely used and straightforward to implement. The novelty of the attack that we propose in this paper is (i) the class of methods that the attack targets and (ii) the way it targets them. Optimization approaches other than PGD could doubtless be used for the same effect.
% note: for consistency, not for any particularly strong reason.

The threat model for the {\emph{token pushing (TP) attack}} that we describe below, as well as a general framework for adversarial attacks on CBIMs, can be found in Section \ref{subsect-threat-model}. At a high-level though, the attack has targeted and untargeted version. 

\textbf{Untargeted attack:} The adversary attempts to modify exemplars for concept $C$ so as to maximally change the interpretation of input with respect to $C$.

\textbf{Targeted attack:} The adversary attempts to modify exemplars for concept $C$ so that interpretations of any input with respect to $C$ now resemble interpretations with respect to a different {\emph{target concept}} $C'$.

%In the positive version the adversary tries to make the interpretation of input $x$ with respect to one concept $C$ align with the interpretation with respect to another (target) concept $C'$. For example, an adversary might seek to make the importance of the concept `bubbles' resemble the importance of the concept `striped' in a model that classifies zebras. Since `striped' is presumably an important concept, the attack would make `bubbly' also appear to be an important concept, which would reduce confidence in a model's predictions. On the other hand, the negative version of the attack simply tries to maximally change the interpretation of input. This might 

The basic idea is simple; we find perturbations to alter a model's internal representation of the concept tokens $P_C = \{x^C_i\}_i$. Using the notation developed in \ref{subsect-threat-model}, let $f: X \rightarrow \mathbb{R}^d$ be a copy of the defender's model or a surrogate. Let $\ell$ be the layer of $f$ that the attack is optimized for.

In the untargeted version, perturbations $\Delta^\ell = \{\delta_i^\ell\}_i$ added to each element in $P_C$ shift their hidden representations in layer $\ell$ so that they no longer correlate with concept $C$. In order to find a point that can guide this shift, the adversary chooses some collection of images that are unrelated to $C$, $U_C := \{x_i^U\}_i$. The adversary calculates the centroid of $f_\ell(U_C)$, which we denote by $\mu_U$. This will serve as a representative of ``unrelatedness'' to $C$ in layer $\ell$. Then for each $x^C_i \in P_C$, the adversary uses PGD to compute
\begin{equation}\label{hiddenstate_untargeted}
    \delta^\ell_i := \argmin_{\|\delta^\ell\|_{\infty} \leq \epsilon}||f_\ell (x_i^C+\delta^\ell)- \mu_U||,
\end{equation}
where $\epsilon > 0$ is chosen based on how visible the attack is allowed to be.  The targeted version of the attack is analogous except that the adversary chooses a target concept $C'$, calculates the centroid $\mu_{C'}$ of $f_\ell(P_{C'})$, and then optimizes for
\begin{equation}\label{hiddenstate_targeted}
    \delta_i^\ell := \argmin_{\|\delta^\ell\|_{\infty} \leq \epsilon}||f_\ell (x_i^C+\delta^\ell)- \mu_{C'}||.
\end{equation}
Both \eqref{hiddenstate_untargeted} and \eqref{hiddenstate_targeted} are related to the hidden layer attacks described in \citep{wang2018great, inkawhich2019feature}. A schematic of the targeted TP attack can be found in Figure \ref{fig-schematic-of-attack}.

In Section \ref{sect-experiments}, we show that in spite of the fact that neither \eqref{hiddenstate_untargeted} nor \eqref{hiddenstate_targeted} is the primary optimization objective of either TCAV or FFV, the TP attack is still effective when applied to either method. In fact, objective functions \eqref{hiddenstate_untargeted} and \eqref{hiddenstate_targeted} make the TP attack more flexible since they act against the underlying linear probe mechanism common to many CBIMs. This means that the adversary does not need to know the specific CBIM that the defender is using in order for the method to have a high probability of success.

\section{Experiments}
\label{sect-experiments}
To better understand the effectiveness of the methods proposed in Section \ref{sect-tokens}, we apply the TP attack to TCAV and FFV. For both TCAV and FFV we focus on InceptionV1 weights trained on ImageNet-1k \citep{deng2009imagenet} from Torchvision \cite{marcel2010torchvision}. We apply our attack to interpretation input from ImageNet and Caltech-UCSD Birds 200 (CUB) \cite{WelinderEtal2010}. The token sets that we use to capture concepts for ImageNet input come from ImageNet itself and the Describable Textures Dataset (DTD) \citep{cimpoi14describing}. The tokens that we use for CUB input come from the attribute metadata associated with that dataset. We perform all PGD attacks with $\epsilon = 8/255$ and 20 steps. To train a CAV, we use a linear classifier trained via stochastic gradient descent and $\ell_2$-regularization. See Section \ref{appendix-exeriment-details} in the Appendix for other experimental details. Examples of perturbed tokens can be found in Figure \ref{fig:example_of_perturbation} in the Appendix. The results we describe in the first part of this section focus on the white-box setting where the adversary knows the defender's model. In Section \ref{sect-transferability} we show that our attacks are also effective in the black-box setting.

\begin{figure}%
\centering
\label{fig:ImageNettarget}%
\includegraphics[height=1.4in]{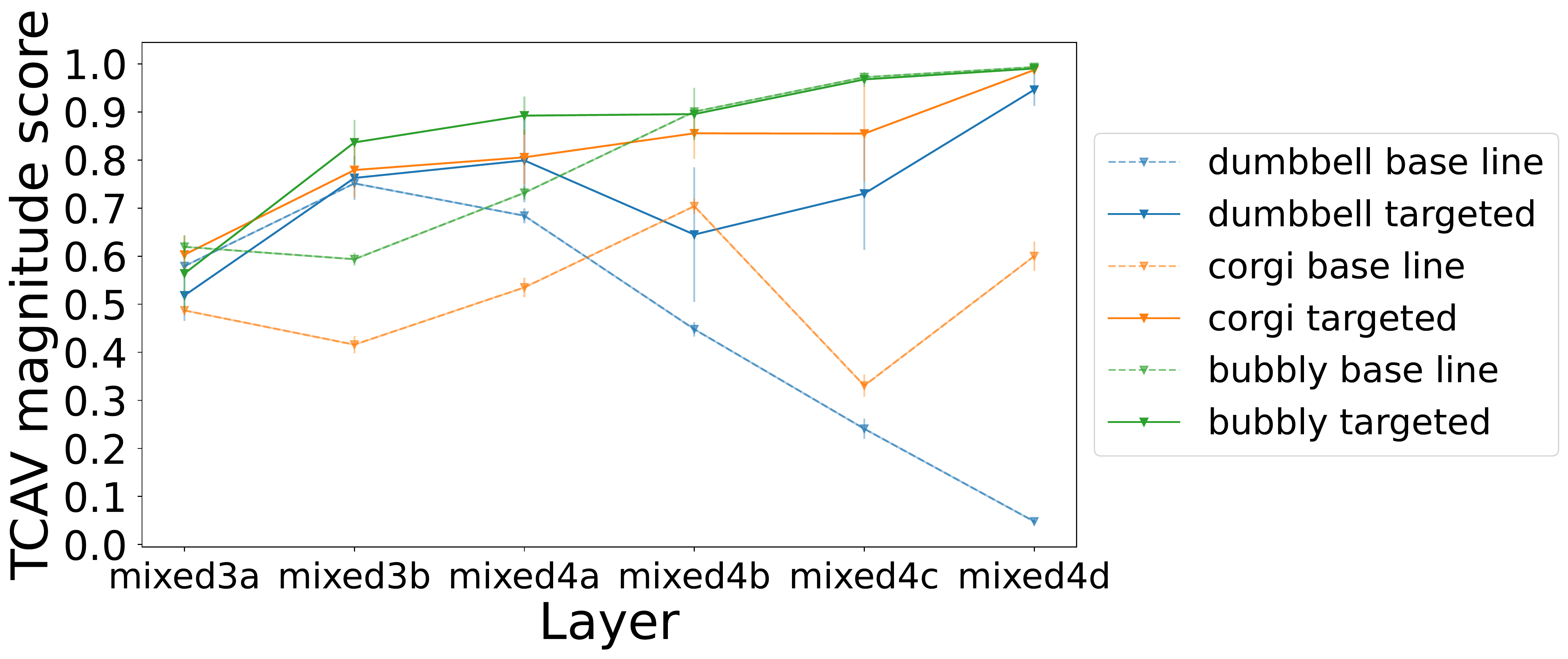}%
\caption{The targeted TP attack, perturbing three classes (dumbbell and corgi from ImageNet, bubbly from DTD) towards the centroid of the honeycombed DTD concept for the layer. TCAV magnitude scores are with respect to the honeycomb ImageNet class.}
\end{figure}

% \begin{figure}%
% \centering
% \label{fig:ImageNettarget}%
% \begin{subfigure}{.5\textwidth}
%   \centering
%   %  trim={<left> <lower> <right> <upper>}
%   \includegraphics[width=0.74\linewidth, trim={0 0 5cm 0},clip]{resnet18_transfer_vanilla_plot.png}
%   \caption{Before attack.}
%   \label{fig:sub1}
% \end{subfigure}%
% \begin{subfigure}{.5\textwidth}
%   \centering
%   \includegraphics[width=\linewidth]{resnet18_transfer_attack_plot.png}
%   \caption{After attack.}
%   \label{fig:sub2}
% \end{subfigure}
% \caption{TCAV sensitivity scores for the zebra class with the stripe images for a ResNet-18 \citep{he2016deep} trained on ImageNet-1K. Attack uses perturbations made on the stripe concept for InceptionV1. The attack is successful on the first residual block. Results shown are all significant, as tested by a two-sided t test.}
% \label{fig:test}
% \end{figure}

\subsection{TP Attacks on TCAV}

To test the untargeted TP attack against TCAV, we choose concept/class pairs with straightforward associations. For example  `striped'/`zebra'. The goal of the attack is to change the interpretation so that a concept that is actually significant to a model, no longer appears so. For example, the perturbation may cause TCAV to indicate that `striped' is not a significant concept for the class `zebra'.  We provide a full list of concept/class pairs in Table \ref{table-untargeted-imagenet-concept-class-pairs} of the Appendix. We perform the same experiment for all concept/class pairs, but for simplicity explain the procedure with the `striped'/`zebra' concept/class pair. We select $70$ non-overlapping sets of $50$ randomly chosen images from ImageNet to be $\{N_{\text{striped}}^i\}$. This same $\{N_{\text{striped}}^i\}$ will be used for all concept/class pairs. We fix a set of unrelated images $U_{\text{striped}}$ of size $1000$ that are also randomly sampled from ImageNet. Finally, we choose random sets of $40$ images from the class `striped', $P_{\mathrm{striped}}$, from DTD. The interpretation input, $D_{\text{zebra}}$, is a collection of images which the model predicts as belonging to the class `zebra'.

For each layer $\ell$ of the model we run the TP attack against $P_{\mathrm{striped}}$ to generate perturbed tokens $\hat{P}_{\mathrm{striped}}^\ell$. For each of the resulting pairs $(P_{\mathrm{striped}},\hat{P}_{\mathrm{striped}}^\ell)$ and each layer $\ell'$ of the model, we then apply TCAV $70$ times (once for each $N_C^i$), calculating the difference in magnitude TCAV score when using $\hat{P}_{\mathrm{striped}}^
\ell$ instead of $P_{\mathrm{striped}}$. Thus in effect, we not only explore the case where the TP attack targets the same model layer that the interpretability method is being used to analyze ($\ell = \ell'$), we also investigate the case where these are different ($\ell \neq \ell'$).

In the targeted case, we focus on concept/class pairs that would not be expected to have any association. For example, class `honeycomb' and concept `Pembroke Welsh corgi'. Then we choose target concepts that would be assumed to be important to the class. For example, we might attack tokens for the concept `Pembroke Welsh corgi' so that it looks like it has the same significance to the ImageNet class `honeycomb' as the DTD texture `honeycombed'. Thus we make it appear that `Pembroke Welsh corgi' is an important concept when the model predicts something is a honeycomb.%Other than the substitution of $U_C$ for $P_{C'}$, the set-up for the positive experiments is identical to the set-up for the negative experiments.

\subsection{TP Attacks on FFV}

%We evaluate the TP attack on FFV by performing feature visualizations for InceptionV1 on every channel neuron for the layers mixed3a, mixed3b, mixed4a, and mixed4b. We optimize for visualizations using objective \eqref{eq:ffv} and then compare these visualizations based on whether they were generated with: clean concept images $P_C$, concept images with Gaussian noise, or concept images to which either a targeted or untargeted TP attack has been applied. We give an example of FFV output before and after an untargeted attack in layer mixed4d in Figure \ref{fig-ffv-visualizations}.

\begin{figure}
\centering
%  trim={<left> <lower> <right> <upper>}
% \includegraphics[width=5.5 in, trim={1cm 0 3cm 1cm},clip]{FVfacetattackexamples.pdf}
% \includegraphics[width= 2.5 in, trim={1cm 1.7cm 2cm 1cm},clip]{fid_run.pdf}
\includegraphics[width= 1 \columnwidth, trim={0cm 1cm 2cm 1cm},clip]{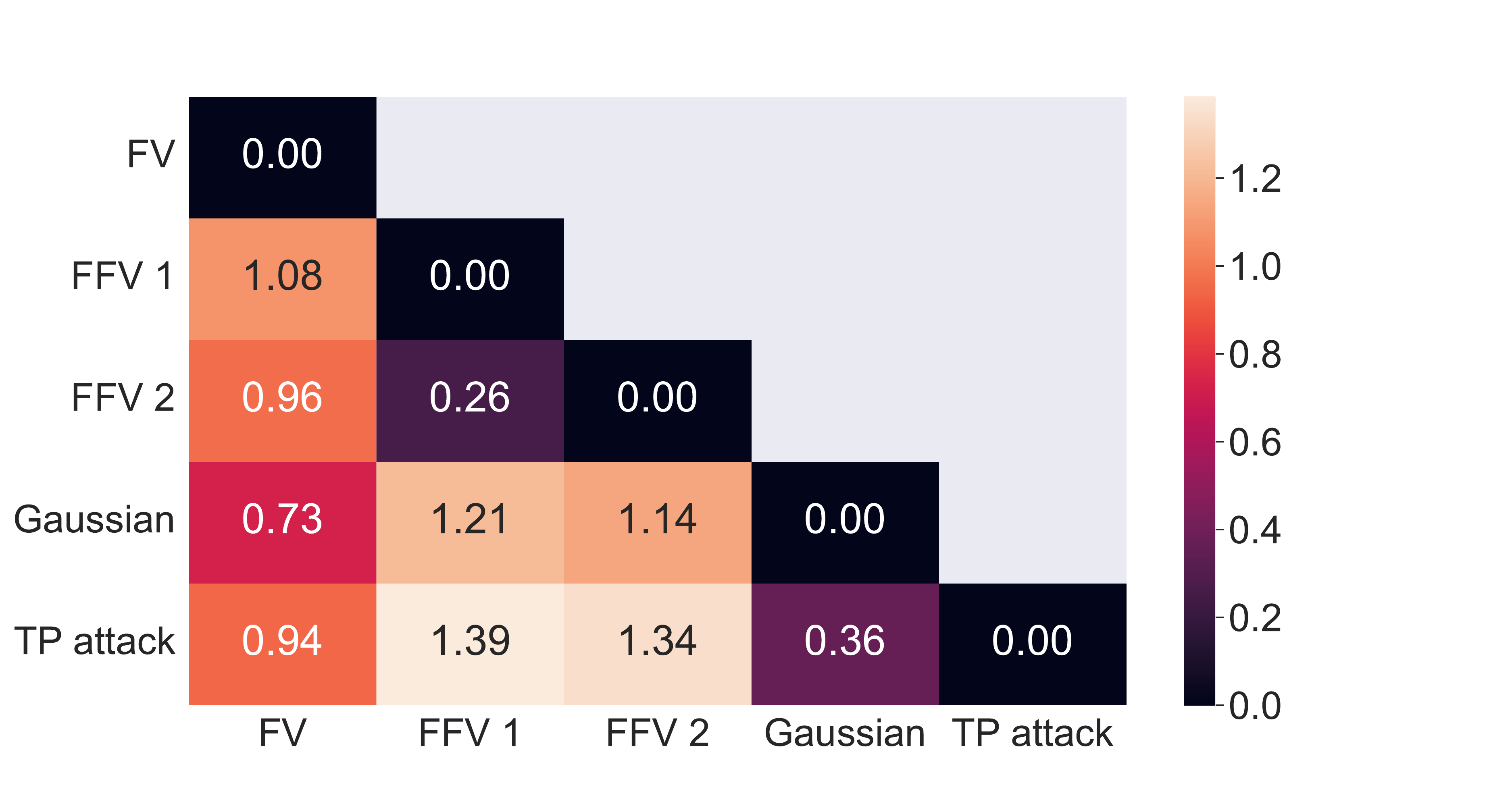}
\caption{Average Fréchet Inception distances between feature visualizations generated from InceptionV1 in different ways: using only the channel term from \eqref{eq:ffv} (\textbf{FV}), two separate runs of FFV with different sets of positive and negative concept images (\textbf{FFV 1} and \textbf{FFV2}), with Gaussian noise added to the positive concept images (\textbf{Gaussian}), and with the token pushing attack applied (\textbf{TP attack}). Targeted layers are mixed3a, mixed3b, mixed4a, and mixed4b.} %The TP attack results in visualizations that are more distinct from the FFV runs in terms of FID than any other method.}
\label{fid} 
\end{figure}

We evaluate the TP attack on FFV by performing feature visualizations for InceptionV1 on every channel neuron for the layers mixed3a, mixed3b, mixed4a, and mixed4b using (1) FV: the channel objective only (i.e., using only the first term in equation \ref{eq:ffv}), (2) FFV1 and FFV2: two different groups of concept images for $P_C$ (`striped') and $N_C$, (3) Gaussian: concept images to which Gaussian noise has been added for $P_C$, and (4) TP attack: concept images to which a TP attack has been applied targeting layer mixed3b for $P_C$. We then compare these visualizations using a variant of the Fréchet Inception Distance (FID) \citep{heusel2017gans} to measure the perceptual distance. A successful attack should significantly change this distance since the visualizations will no longer be optimized towards the ``same'' concept. The FID score is calculated across neurons in all the layers listed above, though our attack only targets mixed3b. We use a PyTorch implementation of FID \citep{Seitzer2020FID} and use the second block of InceptionV3 as the visual similarity encoder (due to the smaller dataset size).

% \drb{and perturbations for more layers if we have time; also, cite future table here.}.

% \begin{figure}
% \centering
% \begin{subfigure}{.5\textwidth}
%   \centering
%   %  trim={<left> <lower> <right> <upper>}
%   \includegraphics[width=0.74\linewidth, trim={0 0 5cm 0},clip]{resnet18_transfer_vanilla_plot.png}
%   \caption{Before attack.}
%   \label{fig:sub1}
% \end{subfigure}%
% \begin{subfigure}{.5\textwidth}
%   \centering
%   \includegraphics[width=\linewidth]{resnet18_transfer_attack_plot.png}
%   \caption{After attack.}
%   \label{fig:sub2}
% \end{subfigure}
% \caption{TCAV sensitivity scores for the zebra class with the stripe images for a ResNet-18 \citep{he2016deep} trained on ImageNet-1K. Attack uses perturbations made on the stripe concept for InceptionV1. The attack is successful on the first residual block. Results shown are all significant, as tested by a two-sided t test.}
% \label{fig:test}
% \end{figure}
\begin{figure}
\centering
\includegraphics[width=1\columnwidth]{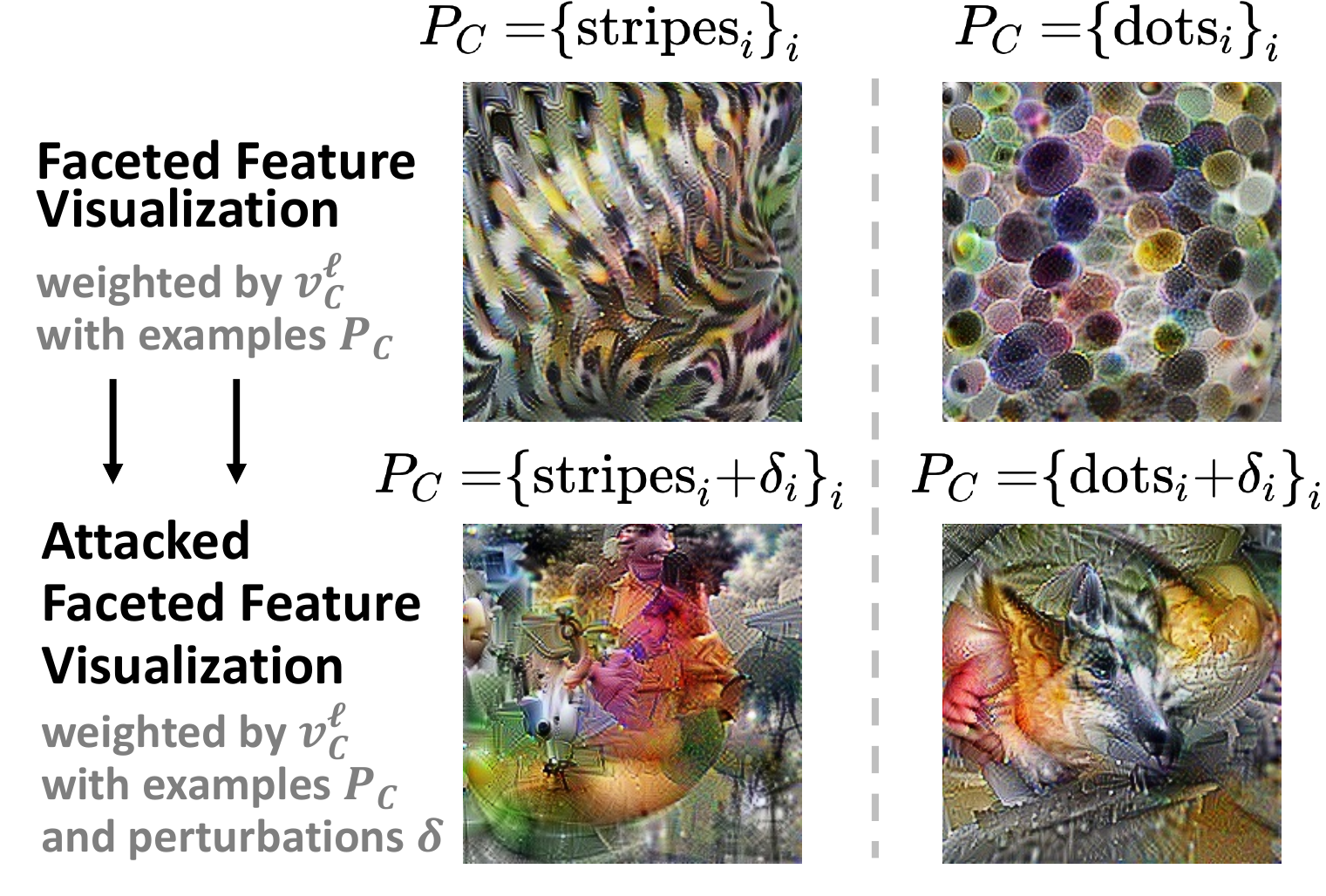}
\caption{A faceted feature visualization of the same neuron (channel 9 on InceptionV1, layer mixed4d) for `striped' and `dots' facets, (first row), and the FFV after a TP attack (second row). While visualizations in the first row reflect the concept priors, the visualizations in the second row do not (indicating the attack was successful).}
\label{fig-ffv-visualizations}
\centering
\label{fig1} 
\end{figure}

 \begin{figure*}
\centering
\includegraphics[width= 1.9\columnwidth]{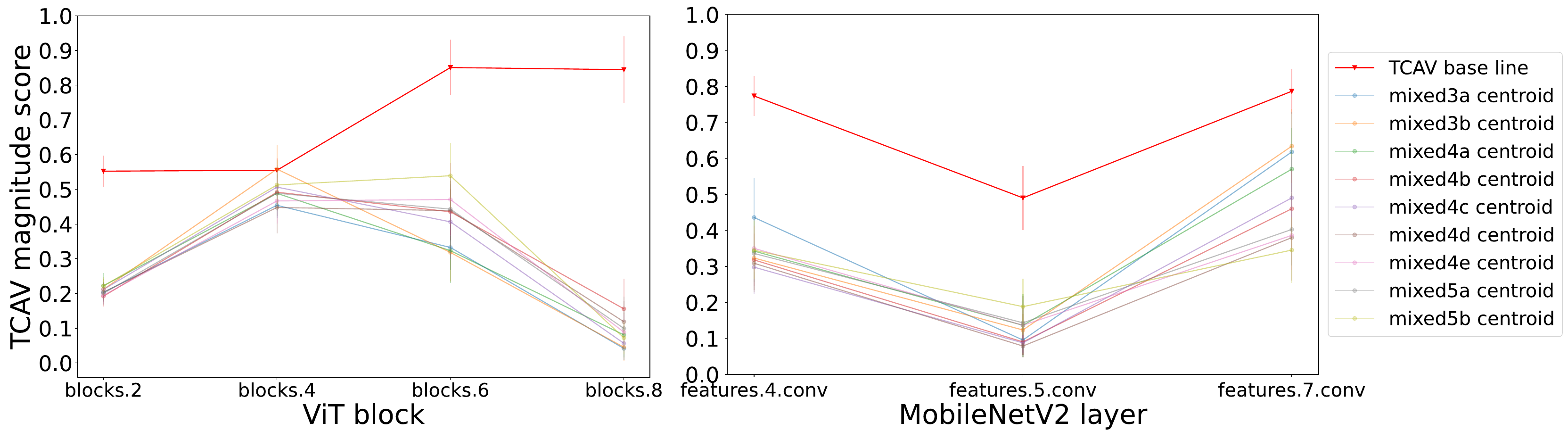}
\caption{TCAV sensitivity scores for the zebra class with the stripe images for a MobileNetV2 (left) \citep{sandler2018mobilenetv2} and a Vision Transformer (right) \citep{dosovitskiy2020image} trained on ImageNet-1K. The attacks use perturbations made on the stripe concept images for InceptionV1 using centroids for different hidden layers (different colored curves). All layers/blocks shown are sensitive to the stripe concept before the attack, and are not sensitive after the attack.}
\label{resnet18} 
\end{figure*}

\section{Results}

Plots of raw TCAV magnitude scores over model layer for both clean concept tokens (dotted lines) and the attacked concept tokens (solid lines) can be found in Figure \ref{fig:CUB}. In the plot on the left the defender's model is a ResNet-18 trained on Caltech-UCSD Birds 200 with TCAV magnitude scores calculated with respect to fixed class `brewer blackbird' and varying concepts. In the plot on the right the defender's model is an InceptionV1 and the fixed concept `scaly' is evaluated with respect to various snake classes. We see that in both cases, our attack results in significant changes in TCAV magnitude scores, meaning that the interpretation of the class in terms of the concept is significantly different before and after the attack (the goal of the untargeted attack). For example, in the right plot in Figure \ref{fig:CUB} we see that the importance of the `scaly' concept for all the snake classes decreases significantly which would signify, to a user who is unaware of the attack, that `scaly' is not important to the model's prediction of snake classes. 

We note that while TP attacks are generally effective, this effectiveness depends on the class, concept, and layer. We see that the attacked `scaly' tokens result in TCAV magnitude scores that are only marginally lower than the baseline at layer `mixed3b' for the class `green snake', whereas the score is much lower at layer `mixed4a'. On all the plots we include $95\%$ confidence intervals for each layer based on the $70$ different $N_C^i$ sets. The point of this is to verify that the result does not depend on having the ``right'' negative examples and to provide evidence that our results are statistically significant.

Figure \ref{fig:ImageNettarget} shows a plot for the targeted TP attack on TCAV. The model being interpreted is an InceptionV1, the concepts being attacked are dumbbell, corgi, and bubbly, and the target class is honeycombed. We see that after the targeted attack, the TCAV scores for all three concepts are higher than their baseline scores, suggesting that the corgi, dumbbell, and bubbly concepts are important in all layers of the model for honeycombed classification.

For FFV, we observe TP attack effectiveness from the differences between the visualizations FFV produces when given a clean concept set $P_C$ and the visualizations FFV produces when given an attacked concept set $\hat{P}_C$. We give two such examples separately using the `striped' and `dotted' concept sets in Figure \ref{fig-ffv-visualizations}. We note that while the first row appears to look like the labeled concept, the second row of attacked visualizations do not appear related to the concept. For quantitative measurements, Figure \ref{fid} gives the average FID between visualizations produced in different ways. We note that while the FID scores between the separate clean FFV runs is $0.26$, the FID score between the TP attack and the clean FFV runs are $1.39$ and  $1.34$. The larger FID scores suggest that the TP attack modifies the FFV output significantly more than the usual variation between runs. %We do note however, that a significant difference in visualization occurs when noise is injected into positive tokens (as shown by the fact that FID for clean FFV runs and runs with Gaussian noise is $1.21$ and $1.14$). 
This, along with visualizations such as \ref{fig-ffv-visualizations}, suggest that a TP attack can drastically change the semantic meaning associated with the feature visualizations produced by FFV.
% It's worth noting that adding Gaussian noise to the striped concept set also leads to a higher FID score than those between the FFV runs.

Finally, we find that both the TCAV magnitudes (Table \ref{table:stripe}) and the FFV FID scores (Figure \ref{fid}) are susceptible to Gaussian noise added to the concept set. This suggests that, even independent of adversarial attacks, CBIMs are brittle. This brittleness suggests that these methods are also vulnerable to natural distribution shifts in data, e.g., between the concept set and training images. We see a need for continued research into robust interpretability methods.

%al differences between the concept set and the original model training distribution or the dataset images $D_k$.

\subsection{Transferability to Different Layers and Model Architectures}
 \label{sect-transferability}
We evaluate TP attacks for two kinds of transferability: transferability to methods which target different layers of a model and transferability to different model architectures. We investigated the former by performing attacks developed for one hidden layer $\ell$, on methods targeting a different hidden layer $\ell'$ as described in Section \ref{sect-experiments}. We found that in many cases, TP attack worked comparably well even when the layer being targeted differed from the layer actually used by the interpretability method (see the off-diagonal entries in Figure \ref{table:stripe} in the Appendix).

We also investigate how TP attacks transfer to a defender that is using a different model architecture by applying attacks developed for InceptionV1 to TCAV when it is used to interpret a MobileNetV2 \cite{howard2017mobilenets} and a Vision Transformer \cite{dosovitskiy2020image} models, all trained on ImageNet. %We focus on the concept/class pair `striped'/ `zebra' using the same sets $N_{\mathrm{striped}}$, $U_{\mathrm{striped}}$, and $P_{\mathrm{striped}}$ as were used in experiments in Section \ref{sect-experiments}. 
We compute the TCAV magnitude score for `striped'/`zebra' for the output of the three layers in MobileNetV2 that were sensitive to the stripe concept according to signed TCAV and the output of the even blocks (2, 4, 6, 8, 10) for the ViT. These results are displayed in Figure \ref{resnet18}. We see that other than block 4 of the Vision Transformer, the TCAV magnitude scores decreases significantly even when perturbations are developed on a model architecture different from the one that is being interpreted. 

%We find that the TP attacks targeting any of the $7$ layers of InceptionV1 result in significant decreases in TCAV magnitude score when applied to the first block of ResNet18 and all three layers of the MobileNetV2. The transfer TP attack does not seem to be effective against Block 2 and Block 4 of the Resnet-18.These results point toward TP attack being moderately transferable, especially when TCAV is being applied to earlier layers of the defender's model.

\section{Conclusion}

In this work we show that concept-based interpretability methods, like much of the deep learning modeling pipeline, are vulnerable to adversarial attacks. By introducing subtle changes to the examples of a concept used to drive the interpretation, an adversary can induce different interpretations. The attacks we describe target the linear probe component common to many different concept-based interpretability methods and  thus are general enough to work for multiple methods without modification. We hope that the results of this paper will promote better security practices, not only around the model pipeline itself, but also around the method that is being used to interpret the model.

%that have different interpretability outputs in images and scalar values for image classification models.
%Unfortunately, safety-critical applications, for which interpretability methods are particularly important, are particularly important


\bibliography{icml2022}
\bibliographystyle{icml2022}

%%%%%%%%%%%%%%%%%%%%%%%%%%%%%%%%%%%%%%%%%%%%%%%%%%%%%%%%%%%%%%%%%%%%%%%%%%%%%%%
%%%%%%%%%%%%%%%%%%%%%%%%%%%%%%%%%%%%%%%%%%%%%%%%%%%%%%%%%%%%%%%%%%%%%%%%%%%%%%%
% APPENDIX
%%%%%%%%%%%%%%%%%%%%%%%%%%%%%%%%%%%%%%%%%%%%%%%%%%%%%%%%%%%%%%%%%%%%%%%%%%%%%%%
%%%%%%%%%%%%%%%%%%%%%%%%%%%%%%%%%%%%%%%%%%%%%%%%%%%%%%%%%%%%%%%%%%%%%%%%%%%%%%%
\newpage
\appendix
\onecolumn
\section{Appendix}

\subsection{Ethics Statement}
In this work we highlight the vulnerability of a class of popular interpretability methods to adversarial attack. We chose to explore a threat model wherein the positive tokens for a concept are perturbed. This is of particular concern because (unlike individual input) positive tokens will often be centralized and used collectively by researchers and practitioners many times. Because of this, an attack on this small subset of data may have wide-ranging effects. We hope that by better understanding and communicating this specific threat to interpretability, we can motivate researchers to use best practices around security for interpretability and explainability as they are already encouraged to do for dataset and model creation.

\subsection{Limitations}

In this work we chose two CBIMs to test TP attacks on. While TCAV and FFV do a good job capturing the diversity of such methods, they do not capture their full breadth. In particular, it would be useful to understand how TP attacks behave when they are applied to other types of feature visualization methods, namely those that average over a large number of images or activations \citep{nguyen2016multifaceted, carter2019activation} to build a visualization. Further, while we only consider image classification models, TCAV is agnostic to modality. Evaluating CBIM brittleness in other critical modalities such as NLP would give a more complete picture of these method's vulnerabilities. Finally, the attacks described in this work perturb positive concept tokens. While we argue that in many ways this is the most critical component of the CBIM pipeline (being re-used for many input), to fully understand the attack surfaces of CBIMs, it makes sense to consider attacks on the other inputs to a method: the model itself, negative examples, and the interpretation input.

\subsection{Related work}

\textbf{Interpretability methods:} Because of the size and complexity of modern deep learning architectures, skill is required to extract interpretations of how these models make decisions. Established methods range from those that focus on highlighting the importance of individual input features to those that can give clues to the importance of specific neurons to a particular class. Popular examples of interpretability methods that focus on input feature importance include saliency map methods \citep{selvaraju2017grad, sundararajan2017axiomatic,ribeiro2016model, fong2017interpretable, DBLP:conf/nips/DabkowskiG17, DBLP:conf/iclr/ChangCGD19} which identify those input features (for example, pixels in an image) whose change is most likely to change the network's prediction. 

CBIMs focus on decomposing the hidden layers of deep neural networks with respect to human-understandable concepts. One of the best-known approaches in this direction involves the use of concept activation vectors (CAVs) \citep{kim2018interpretability}. Work that is either related or extends these ideas includes \citep{zhou2018interpretable,graziani2018regression,graziani2019improved, pmlr-v119-koh20a, yeh2020completeness}. 

Feature visualization is a set of interpretability techniques \citep{szegedy2014intriguing, mahendran2015understanding, wei2015understanding, nguyen2016multifaceted} concerned with optimizing model input so that it activates some specific node or set of nodes within the network. 
However, a challenge arises when one tries to analyze `polysemantic neurons' \citep{olah2018the}, neurons that activate for several conceptually distinct ideas. For example, a neuron that fires for both a boat and a cat leg is polysemantic. Interpretability methods have imposed priors to disambiguate neurons by clustering the training images \citep{wei2015understanding, nguyen2016multifaceted} or the hidden layer activations \citep{carter2019activation} and using the average of the cluster as a coarse-grained image prior, parameterizing the feature visualization image with a learned GAN \citep{nguyen2016synthesizing}, or using a diversity term in the feature visualization objective \citep{wei2015understanding, olah2017feature}.

\begin{table*}\centering
\small
\begin{tabular}{@{}rrrrr@{}}\toprule
& \multicolumn{4}{c}{InceptionV1 Layer}\\
\cmidrule{2-5} 
\textbf{Attacks} & \multicolumn{1}{c}{mixed3a} & \multicolumn{1}{c}{mixed3b} & \multicolumn{1}{c}{mixed4a} & \multicolumn{1}{c}{mixed4b} \\ 
\midrule
Baseline TCAV (no attack) & $0.69 \pm 0.02$ & $0.90 \pm 0.01$ & $0.66 \pm 0.03$	& $0.68 \pm 0.04$ \\ 
\midrule
% \textbf{Attacks} \\ 
Gaussian noise & $0.61 \pm 0.02$ & $0.62 \pm 0.02$ & $0.64 \pm 0.03$ & $0.67 \pm 0.04$ \\
\textit{TP attack on} \\ 
Logit & $0.37 \pm 0.02$ & $0.37 \pm 0.03$ & $0.35 \pm 0.02$ & $0.33 \pm 0.03$ \\
mixed3a centroid & $\bm{0.29 \pm 0.05}$ & $0.29 \pm 0.10$ & $0.22 \pm 0.05$ & $0.34 \pm 0.08$ \\
mixed3b centroid & $0.17 \pm 0.05$ & $\bm{0.39 \pm 0.10}$ & $0.19 \pm 0.03$ & $0.37 \pm 0.08$ \\
mixed4a centroid & $0.22 \pm 0.06$ & $0.40 \pm 0.11$ & $\bm{0.32 \pm 0.05}$ & $0.44 \pm 0.08$ \\
mixed4b centroid & $0.27 \pm 0.07$ & $0.32 \pm 0.10$ & $0.33 \pm 0.06$ & $\bm{0.42 \pm 0.08}$ \\
mixed4c centroid & $0.26 \pm 0.08$ & $0.30 \pm 0.09$ & $0.29 \pm 0.05$ & $0.28 \pm 0.08$ \\
mixed4d centroid & $0.28 \pm 0.08$ & $0.30 \pm 0.10$ & $0.25 \pm 0.06$ & $0.18 \pm 0.10$\\
% \\
\bottomrule
\end{tabular}
\caption{The TCAV magnitude score for the zebra class on the `striped' concept, before and after the TP attacks on InceptionV1. The Baseline TCAV row uses the concept sets with no perturbations. The Gaussian noise row applies Gaussian noise to positive tokens. The rows below `TP attack on' indicate the layer that is being targeted by the TP attack. The columns are the InceptionV1 layer that TCAV is being applied to. For all concept/pairs we bold those values where the layer targeted by the TP attack and the layer TCAV is applied to are the same.} \label{table:stripe}

\end{table*}

\textbf{Robustness of interpretability methods:} This is not the first work that has shown that interpretability methods can be brittle. Saliency methods have been shown to produce output maps that appear to point to semantically meaningful content even when they are extracted from untrained models, indicating that these methods may sometimes simply function as edge detectors \citep{adebayo2018local}. While not an interpretability method per se, preliminary work has studied the robustness of Concept Bottleneck Models, an intrinsically interpretable concept-based method, to out-of-distribution data \citep{pmlr-v119-koh20a}. From a more adversarial perspective, a number of works have shown that saliency methods are vulnerable to small perturbations made to either an input image or to the model itself that cause the model to offer radically different interpretations \citep{heo2019fooling,ghorbani2019interpretation,viering2019manipulate,subramanya2019fooling,pmlr-v119-anders20a}; work has looked at methods to make explanations more robust to attack \citep{lakkaraju2020robust}.  On the other hand, this is the first work that shows that CBIMs are also vulnerable to adversarial attack. In particular, since we focus on attacks targeting a component absent from other interpretability methods (concept tokens), there is not a straightforward way of applying the attacks mentioned above within the threat model presented in this paper.

\subsection{A Threat Model for CBIMs}
\label{subsect-threat-model}

We frame the notion of a CBIM abstractly in order to better understand its attack surface. We view such a method as a map that takes (1) a model from family $\mathcal{M}$, (2) positive tokens of the concept that we would like to steer our interpretation (from space $\mathcal{P}$), (3) negative tokens of the concept (from space $\mathcal{N}$), and (4) an {\emph{interpretation input}} which will be the focus of the interpretation (from space $\mathcal{I}$). We call the output of an interpretability method an {\emph{interpretation output}}. An interpretation output might be a single scalar value (as in the case of TCAV), or it may be an image (as in the case of FFV). In all cases, an interpretation output is designed to help the user better understand a model's decision making process. Thus, we can understand a CBIM as a function $T: \mathcal{M} \times \mathcal{P} \times \mathcal{N} \times \mathcal{I} \rightarrow \mathcal{O}$. We note that in the case of TCAV, the interpretation input is a dataset $D_k$ of examples of some class $k$, while the interpretation input of FFV is a specific node position $(i,j,k)$ in the model. 

Since we will only be considering images as input in our experiments, we specify to that setting here. Otherwise, we use the formalism that we developed above. Specifically, we assume there exists an interpretability method $I$, a model $f \in \mathcal{M}$, a set of positive image tokens $P_C = \{x_i^C\}_i \in \mathcal{P}$, a set of negative image tokens $N_C \in \mathcal{N}$, and an interpretation input $I \in \mathcal{I}$. We also assume a function $F:\mathcal{O} \times \mathcal{O} \rightarrow \mathbb{R}$ that quantitatively captures meaningful difference between interpretation output.

{\textbf{Adversary's goal:}} Find perturbations $\{\delta_i\}_i$ to generate a new {\emph{attacked}} positive token set $\hat{P}_C = \{x_i^C + \delta_i\}_i$ to satisfy the following objective functions:
\begin{itemize}[noitemsep]
    \item (Untargeted) maximizes the difference 
    \begin{equation*}
    \argmax_{\{\delta_i\}_i} F(I(f,P_C,N_C,T),I(f,\hat{P}_C,N_C,T)),
    \end{equation*}
    \item (Targeted) minimizes the difference \begin{equation*}
    \argmin_{\{\delta_i\}_i} F(I(f,P_{C'},N_{C'},T),I(f,\hat{P}_C,N_C,T))
    \end{equation*}
    for some second concept $C'$.
\end{itemize} 
In order to avoid detection, $\hat{P}_C$ is subject to the constraint: $\max_{i} ||\delta_i||_{\infty} \leq \epsilon$, for some fixed $\epsilon > 0$.

Informally, in the untargeted setting the adversary tries to maximally alter the way input is interpreted with respect to a concept $C$ (without regard to the direction of the new interpretation), while in the targeted setting the adversary wants the apparent interpretation of output with respect to concept $C$ to actually be as close as possible to the actual interpretation output with respect to some distinct concept $C'$. For example, a untargeted attack on a stop sign classifier might seek to make the concept of `red’ appear unimportant, as a way of reducing trust in the model\footnote{Stop signs are red in North America.}. On the other hand, a targeted attack might seek to change the interpretation with respect to the concept `blue sky’ so that it resembles the true interpretation with respect to `red’. Since `red’ is presumably an important concept to a stop sign classifier, a successful attack of this type would cause `blue sky’ to also seem like an important concept to the model. Since the background weather should not be an important concept for the task of identifying a stop sign, this could also cast doubt on the model’s reliability.

{\textbf{Adversary knowledge and capabilities:}} 
In this paper we assume that the adversary has read and write access to the tokens $P_C$ either before or after they have been collected. We also assume that the adversary has access to at least a surrogate of the model that is being interpreted. We discuss transferability of the attack in Section \ref{sect-transferability}.

The adversary's goal is framed in terms of a function $F$ that depends on the specific interpretability method. We show that TP attacks, which we propose below, work without modification for a range of $F$, including those for TCAV and FFV, by optimizing for an objective function that disrupts the fundamental mechanism underlying most CBIMs. As noted in the introduction, we centered our threat model around the positive tokens critical to CBIMs that, once perturbed, can cause persistent misinterpretation across numerous inputs. In contrast, a perturbation of an individual input image alone affects only the interpretation associated to that input.

\subsection{Further Experimental Details}
\label{appendix-exeriment-details}
To run TCAV, FFV, and our attacks, we use PyTorch with an NVIDIA Tesla T4 GPU provided with Google Colab Pro as well as a single NVIDIA Tesla P100 GPU. We use the Captum \citep{kokhlikyan2020captum} implementation of TCAV with a linear classifier trained via stochastic gradient descent and $\ell_2$-regularization. For the Faceted Feature Visualizations, we start with random noise and parameterize the image Fourier basis \citep{olah2017feature}. We use random scaling, rotation, color, and shift transformations.

The concept/class pairs that we used for untargeted attacks on ImageNet classes can be found in Table \ref{table-untargeted-imagenet-concept-class-pairs}. For CUB we used concepts taken from the CUB metadata attributes: `has bill shape all purpose',
`has bill shape needle', `has bill shape spatulate',
`has primary color red'. We pair each of these with each class in a size $70$ subset of CUB classes. The concept/class/target concept triplets that we used for targeted attacks on ImageNet input can be found in Table \ref{table-targeted-imagenet-concept-class-pairs}.

In our transferability experiments in Section \ref{sect-transferability}, for all models we use Torchvision pretrained weights \cite{marcel2010torchvision}, except for the ViT which uses the implementation and pretrained weights found in \cite{rw2019timm}.

\begin{table}[t]
\caption{The concept/class pairs used for the untargeted ImageNet experiments described in Section \ref{sect-experiments}. Concept examples are either taken from ImageNet itself or DTD.} 
\label{table-untargeted-imagenet-concept-class-pairs}
\begin{center}
\begin{tabular}{ll}
%\multicolumn{4}{c}\\\hline\hline
%\multicolumn{4}{c}{\textsc{Fiberwise generation performance ($\mathcal W_1$ distance)}}\\\hline\hline
             \textbf{Concept} & \textbf{Class}
\\ \hline 
Honey & Honeycombed\\
Zebra & Striped\\
% Honeycomb & Honeycombed\\
Green snake & Scaly\\
Hognose snake & Scaly\\
Water snake & Scaly\\
King snake & Scaly\\
% Saltshaker & Dining table\\
% Dalmatian & Black-footed ferret\\
% Persian cat & Cougar\\
% Flat-coated retriever & Newfoundland\\
% Digital clock & Wall clock\\
% Wall clock & Digital clock\\
% Dining table & Table lamp\\
\end{tabular}
\end{center}
\end{table}

\begin{table}[t]
\caption{The concept/class/target class triplets used for the targeted ImageNet experiments described in Section \ref{sect-experiments}. Concept examples are either taken from ImageNet itself or DTD.} 
\label{table-targeted-imagenet-concept-class-pairs}
\begin{center}
\begin{tabular}{lll}
%\multicolumn{4}{c}\\\hline\hline
%\multicolumn{4}{c}{\textsc{Fiberwise generation performance ($\mathcal W_1$ distance)}}\\\hline\hline
             \textbf{Concept} & \textbf{Class} & \textbf{Target class}
\\ \hline 
Bubbly & Honeycomb & Honeycombed\\
Dumbbell & Honeycomb & Honeycombed\\
Corgi & Honeycomb & Honeycombed\\\end{tabular}
\end{center}
\end{table}

\subsection{TP Attack on Relative TCAV}
\label{sect-relative-tcav}

\begin{figure}
\centering
\includegraphics[width=1\columnwidth]{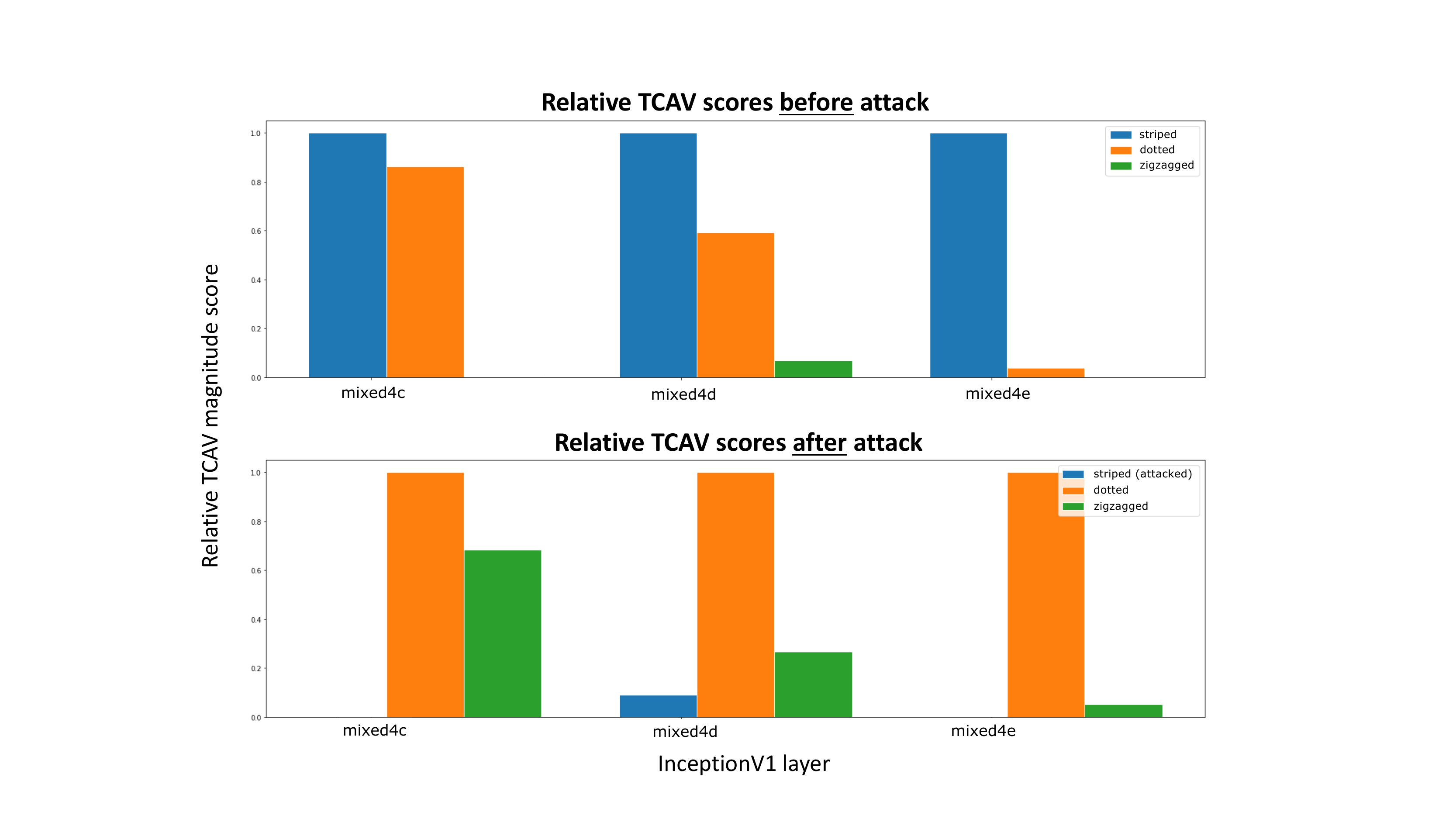}
\caption{Relative TCAV magnitude scores before (top) and after (bottom) the TP attack on the `striped' concept images. Note that the `striped' concept goes from being a relatively more important concept (before attack) to an unimportant concept (relative to concepts `zigzagged' and `dotted').}
\label{fig:rel} 
\end{figure}

As mentioned in Section \ref{sect-TCAV} the relative TCAV score aims to measure the importance of one concept relative to another. We show that the TP attack is also effective against this variant of TCAV. We again focus on input class `zebra'. It would be expected that the importance of the concept of `striped' would be high relative to the concepts of `zigzagged' or `dotted' and indeed we see this experimentally for an InceptionV1 model in the top plot of Figure \ref{fig:rel}. On the other hand, after applying an untargeted TP attack to `striped', we see that `dotted' becomes vastly more important than `striped' in all cases (as seen in the bottom plot of Figure \ref{fig:rel}), while `zigzagged' becomes significantly more important than `striped' in layer mixed4c and slightly more important in layers mixed4d and mixed4e. 

%\subsection{Striped images before and after TP attack}
\begin{figure}[!ht]
\centering
\includegraphics[width=4 in]{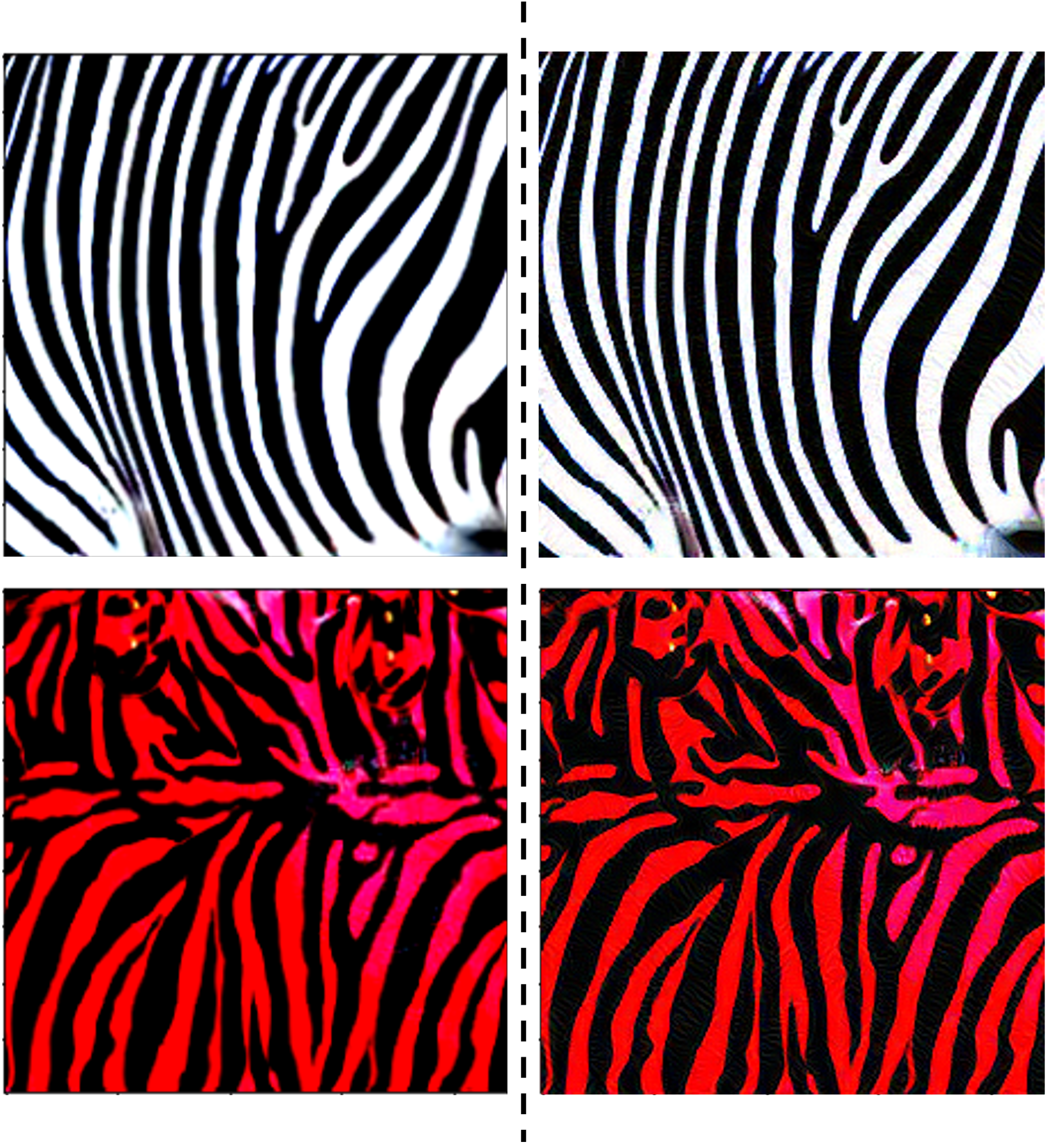}
\caption{Example of `striped' concept images before (left) and after (right) an untargeted TP attack using $\epsilon = 8/255$ and 20 iterations of PGD. The perturbation shown targets InceptionV1 layer mixed3a.}
\label{fig:example_of_perturbation} 
\end{figure}

% \begin{figure}[!ht]
% \includegraphics[width=6 in]{tcav_score_before_and_after.pdf}
% \caption{Relative TCAV scores between the striped before (top) and after (bottom) the attack, zig zag, dotted concept sets.}
% \centering
% \label{fig2} 
% \end{figure}

\subsection{Can Attacked CAVs be Detected with DeepDream?}
\begin{figure}[!ht]
\centering
\includegraphics[width=5 in]{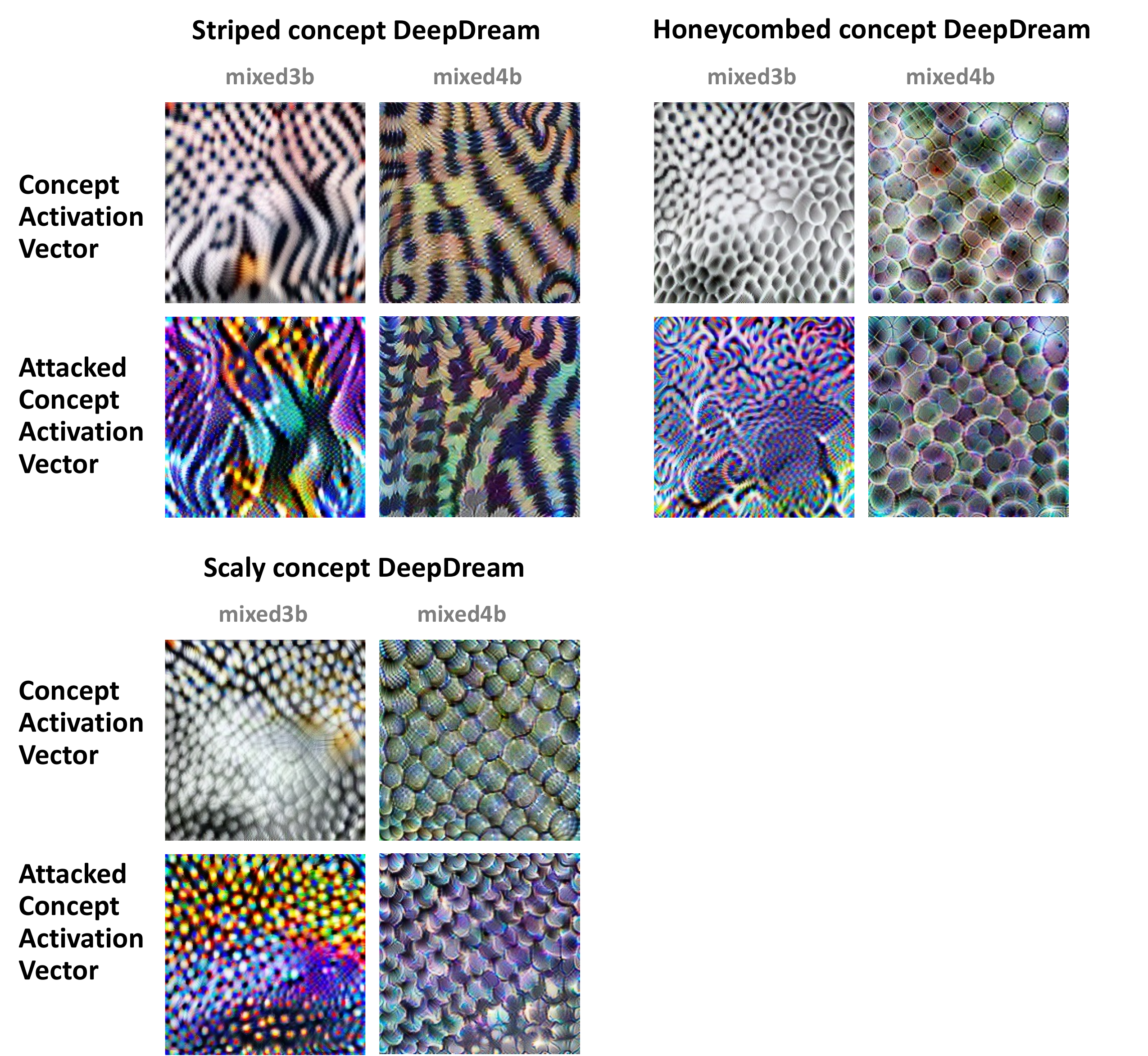}
\caption{Empirical Deepdream \cite{mordvintsev2015deepdream} visualizations for CAVs computed from the original concept sets $P^\ell_C$ (top row of each grid) and the attacked concept sets $\hat{P}^\ell_C$ (bottom row of each grid). Columns within the grids correspond to CAVs in different layers of the model. Each grid corresponds to a different concept (`striped', `honeycombed', `scaly'). For the attacked concept sets, the TP attack targets hidden layer mixed4d of InceptionV1. Note that except for some strange coloring, the visualizations still resemble the initial concept, suggesting that DeepDream may not be an effective tool for identifying attacked tokens.}
\label{fig:deepdream} 
\end{figure}

Could a perturbed concept set $\hat{P}_C^\ell$ itself be identified as corrupted through visualization? Might this be a possible defense against TP attacks? To investigate this, we applied Empirical DeepDream to CAVs to which an untargeted TP attack had been applied \cite{mordvintsev2015deepdream}. These are shown in Figure \ref{fig:deepdream} where we use DeepDream to visualize a CAV before and after the TP attack. We consider CAVs for the hidden layers mixed3b and mixed4b of InceptionV1. We use images from the `striped', `honeycombed', and `scaly' concept sets, and use a TP attack aimed at the hidden layer mixed4d. We use cosine similarity \citep{carter2019activation} for the feature visualization objective and the same Fourier parameterization and transformations we used for the FFV.

We note that the visualizations for the attacked CAV tend to qualitatively resemble those of the CAV without the attack, albeit with unnatural hue and colors. It has been proposed that DeepDream can confirm that CAVs represent the concept of images \citep{kim2018interpretability}. The small experiments we describe here suggest this approach is not an effective defense against TP attacks since attacked CAV tend to visually resemble unattacked CAVs.%Given the success of our adversarial attack, using DeepDream as a qualitative concept check for a CAV may therefore be misleading and provides evidence for another level of imperceptibility of the TP attack.

%\citep{santurkar2021editing}
%%%%%%%%%%%%%%%%%%%%%%%%%%%%%%%%%%%%%%%%%%%%%%%%%%%%%%%%%%%%%%%%%%%%%%%%%%%%%%%
%%%%%%%%%%%%%%%%%%%%%%%%%%%%%%%%%%%%%%%%%%%%%%%%%%%%%%%%%%%%%%%%%%%%%%%%%%%%%%%

\end{document}